\documentclass[letterpaper, 10 pt, conference]{ieeeconf}  

\IEEEoverridecommandlockouts                              

\usepackage{amsthm}
\usepackage{graphics} 
\usepackage{epsfig} 
\usepackage{mathptmx} 
\usepackage{times} 
\usepackage{amsmath} 
\usepackage{amssymb}  

\usepackage{biblatex}
\usepackage{bm}
\usepackage{xcolor}
\usepackage{graphicx}
\usepackage[ruled,linesnumbered]{algorithm2e}
\usepackage[noend]{algpseudocode}
\usepackage{booktabs}
\usepackage{siunitx}
\usepackage{hyperref}

\usepackage{caption}
\usepackage{lipsum}
\usepackage{multirow}
\usepackage{multicol}
\usepackage{diagbox} 
\usepackage{subcaption}

\usepackage{newtxtext, newtxmath}

\newtheorem{remark}{\bf Remark}[section]

\title{\LARGE \bf
Distributed Map Classification using Local Observations 
}

\author{Guangyi Liu, Arash Amini, Martin Tak\'a\u{c}, H\'ector Mu\~noz-Avila, and Nader Motee$^{1}$

\thanks{$^{1}$G.L., A.A., and N.M. are with the Department of Mechanical Engineering and Mechanics, Lehigh University, Bethlehem, PA 18015, USA {\tt\small \{gliu,ara416,motee\}@lehigh.edu}. M.T. is with the Department of Industrial and Systems Engineering, Lehigh University {\tt\small \{takac.mt\}@gmail.com}. H.M. is with the Department of Computer Science and Engineering, Lehigh University {\tt\small \{hem4\}@lehigh.edu}.
}
}

\begin{document}

\maketitle

\thispagestyle{plain}
\pagestyle{plain}


\begin{abstract}
We consider the problem of classifying a map using a team of communicating robots. It is assumed that all robots have localized visual sensing capabilities and can exchange their information with neighboring robots. Using a graph decomposition technique, we proposed an offline learning structure that makes every robot capable of communicating with and fusing information from its neighbors to plan its next move towards the most informative parts of the environment for map classification purposes. The main idea is to decompose a given undirected graph into a union of directed star graphs and train robots w.r.t a bounded number of star graphs. This will significantly reduce the computational cost of offline training and makes learning scalable (independent of the number of robots). Our approach is particularly useful for fast map classification in large environments using a large number of communicating robots. We validate the usefulness of our proposed methodology through extensive simulations.  
\end{abstract}


\section{Introduction}

Real-time perception and classification with a multi-robot system have been among the outstanding research areas in robotics for the past decades. The significant challenges lie in developing a multi-robot learning and communication structure with scalability, high accuracy, and low computational complexity. In most real-world applications, robots can only sense their surrounding environments due to sensor capabilities and physical constraints. To reveal more information about the environment, robots may communicate and exchange information about their observation with others. The questions may arise as: What relevant information should they exchange? Should every robot communicate with all other robots or only with its neighbors? Can we localize the learning to reduce the training cost with a large number of robots? Finding answers to these and other similar questions will significantly facilitate tackling perception-based problems over a network of robots. As it is common in almost all networked systems, achieving scalability in design will become the key to having a successful story. 

\begin{figure}[t]
	\centering
	\includegraphics[width=\linewidth]{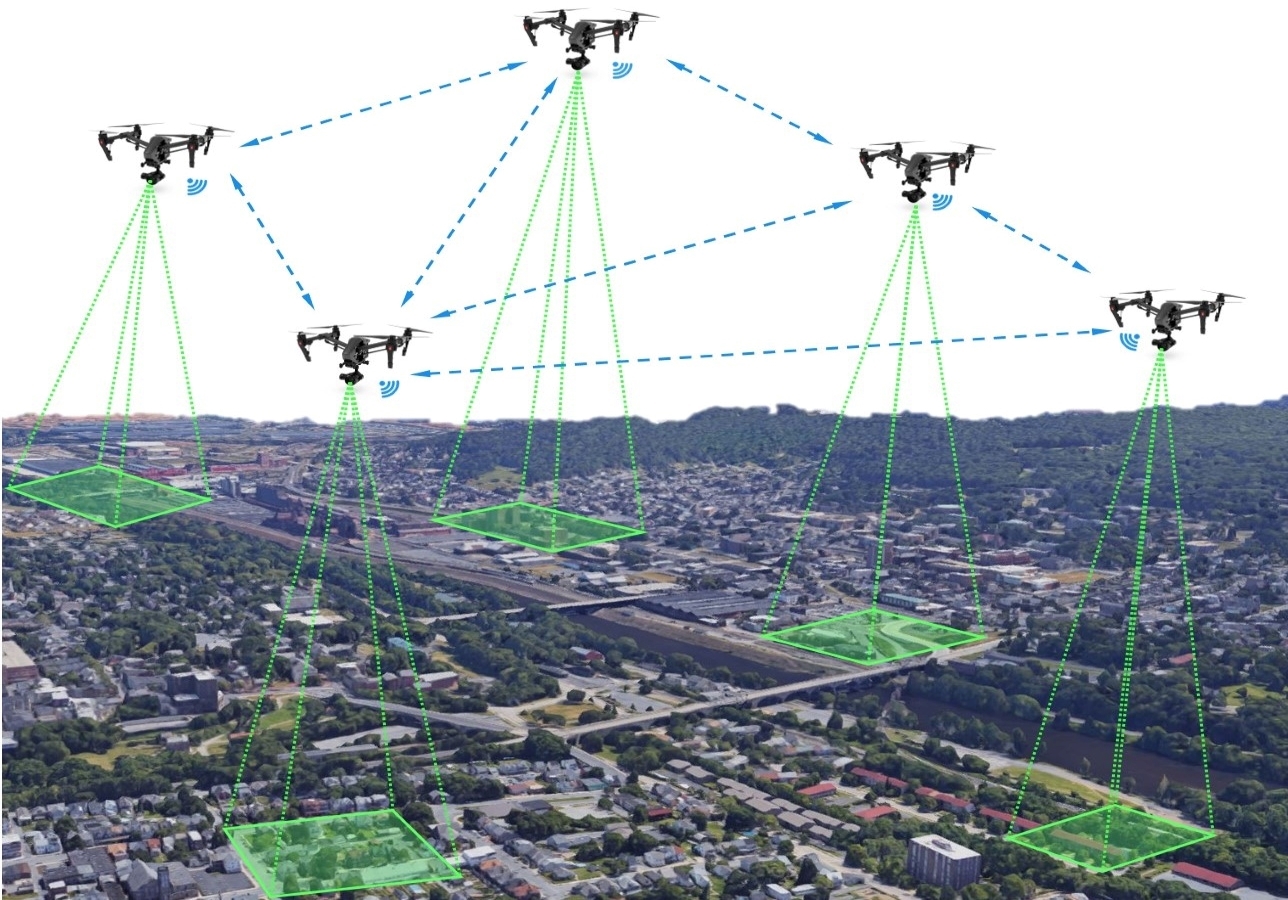}
	\caption{Aerial robots aim to classify a map. Green shaded areas denote localized observations, and blue dashed arrows stand for the communication links.}
	\label{fig:intro}
\end{figure}

In this work, we solve the problem of multi-robot map classification with aerial robots, as shown in Fig.\ref{fig:intro}. Unlike the centralized method, which uses complete information as an input \cite{lund1963map}, we propose a distributed structure that enables a team of robots to classify the target environment as an image. Each robot takes local observations from the environment and communicates with its neighbors to finalize the classification in a distributed manner. 

\begin{figure*}[t]
	\begin{center}
		\makebox[\textwidth]{\includegraphics[width=\textwidth]{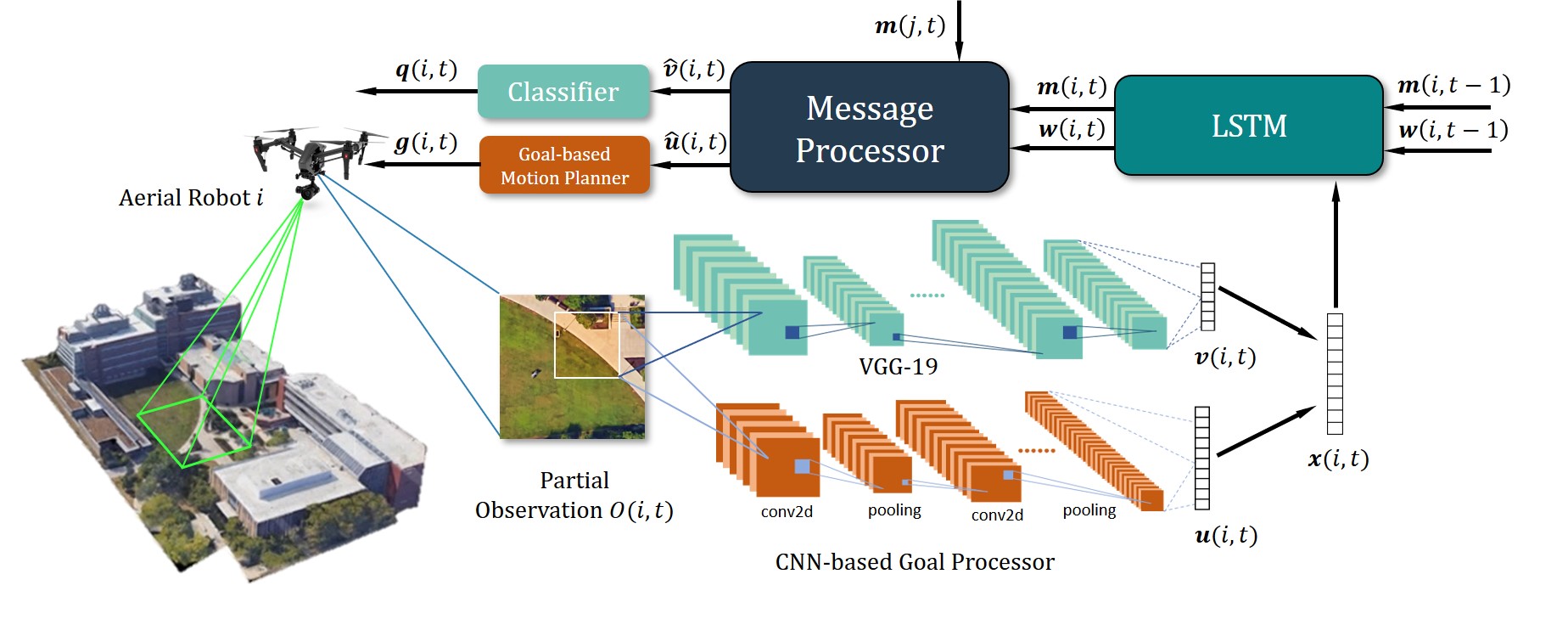}}
		\caption{Distributed map classification architecture, shown for a single robot.}
		\label{fig:diagram}
	\end{center}
\end{figure*}

The limitation on the hardware constrains the communication range, as a result, the network topology is time-varying when robots traverse the environment. The problem for a group of robots to learn effective communication for a fixed topology has been solved in \cite{mousavi2019multi}, but the performance will decrease, and additional training is required if the topology is time-varying. To tackle this problem, we decompose the communication topology with a group of star graphs. Robots then learn to communicate with up to a finite number of neighbors, as a star graph. Such manipulation enables localized and scalable learning for robots since they can always decompose an arbitrary communication topology with star graphs that they have been previously trained with. We refer the details to Section \ref{sec:scalable_learn}. 

Applications of multi-robot map classification include identifying search and rescue regions after natural disasters (e.g., earthquake or tsunami), enhancing existing maps of a region in a city for traffic and crowd control, and constructing soil map in geoscience during or after flooding to prevent and control potential subsequent disasters in a short time \cite{bock2007xv}. In such applications, the classification problem should be solved fast enough using a network of robots that can only observe the environment locally. 

{\it Related Work} Robot's operation involved with learning to classify an image and accomplish a real-time task has been widely studied by \cite{mota2019commonsense, singh2019end}. In real-world scenarios, such functionality and performance of the robot must be preserved when the imagery inputs are unperturbed, or only partially available \cite{sunderhauf2018limits}. The current state-of-the-art approach \cite{karkus2019differentiable} has widely studied to perceive the underlying state with the limited or partial observation by implementation with Reinforcement Learning. Learning the task or skill with a multi-robot system has been proposed by \cite{bucsoniu2010multi,shoham2003multi}. In such a setting, communication is also a vital feature of the multi-robot system. Learning to communicate effectively by using reinforcement learning was studied by \cite{foerster2016learning}. In the case when both communication and information input is constrained, which lands in the area of federated learning, robots can still learn a useful model with the approaches proposed by \cite{konevcny2016federatedX, li2019federated, smith2017federated}, which is essential to our problem since communication is not always available in the real-world scenario.

The recent works about the CNN-LSTM model have inspired us to consider the partial observations as a time-indexed sequence of visual features. \cite{wang2016dimensional} used a combination of regional CNN-LSTM model which divide the text input into several regions and use their weighted contribution for valence-arousal (VA) prediction. Recent works \cite{byeon2015scene, walch2017image, liang2016semantic} have used the CNN-LSTM models for image space-related purposes with promising results. In our work, the CNN-LSTM structures are used among a group of robots and with a time-varying partial input.

\section{Preliminaries and Mathematical Notations}

We assume there are $N$ robots operating in a discrete time horizon $t = \{1,...,T\}$.  The communication topology among robots is represented by a time-varying (undirected) graph $\mathcal{G}(t)=\big( \mathcal{V},\mathcal{E}(t) \big)$, in which $\mathcal{V}$ is the set of vertices (robots), and $\mathcal{E}(t)$ is the set of unweighted communication links. The cardinality of a set is denoted as $|\cdot|$, and $|\mathcal{V}| = N$. The directional edge from node $i$ to node $j$ is denoted by $(i,j)$ for $i,j \in \mathcal{V}$ and $i \neq j$  \cite{van2010graph}. The degree of the node $i$ is shown by $d_i$, it measures the number of connections to other nodes. We denote a node $j$ is an in-neighbor of the node $i \in \mathcal{V}$ at time $t$ if $(j,i) \in \mathcal{E}(t)$. The union of two directional graphs $\mathcal{G}_1=(\mathcal{V}_1, \mathcal{E}_1)$ and $\mathcal{G}_2=(\mathcal{V}_2,\mathcal{E}_2)$ is defined as 
\[ \mathcal{G} = \mathcal{G}_1 \cup \mathcal{G}_2 \]
where $\mathcal{G}=(\mathcal{V},\mathcal{E})$ with $\mathcal{V} = \mathcal{V}_1 \cup \mathcal{V}_2$ and $\mathcal{E} = \mathcal{E}_1 \cup \mathcal{E}_2$. Using this definition, we define and denote an undirected edge between nodes $i$ and $j$ by $\{i,j\} = \{ (i,j)\} \cup \{(j,i)\}$.     

\section{Problem Statement}

The {\it problem} is to classify map of a region with a team of communicating robots. We assume all robots have localized visual sensing capabilities and can exchange their information with their neighbors, as shown in Fig. \ref{fig:intro}. Robots collect local observations as a sequence of images from the environment. Considering the communication range, robots can only establish communication links with each other whenever they are within a certain distance from one another. Robots are expected to traverse in the underlying environment, collect observations, exchange information with others, and finalize the classification with the fused information. Before deployment, robots are trained offline for a given training set of maps. Our {\it objective} is to design a distributed map classification architecture that can successfully solve the proposed problem over time-varying communication graphs using a scalable and localized learning mechanism.


\section{Scalable Learning via Graph Decomposition}     \label{sec:scalable_learn}
In distributed map classification, our objective is to train the robots offline such that they can communicate over arbitrary time-varying graphs and complete the classification task independent of the number of robots. In order words, our goal is to propose a learning mechanism that scales with the number of robots and can handle all possible communication graph topologies.     

\begin{figure}[t]
	\centering
	\includegraphics[width=\linewidth]{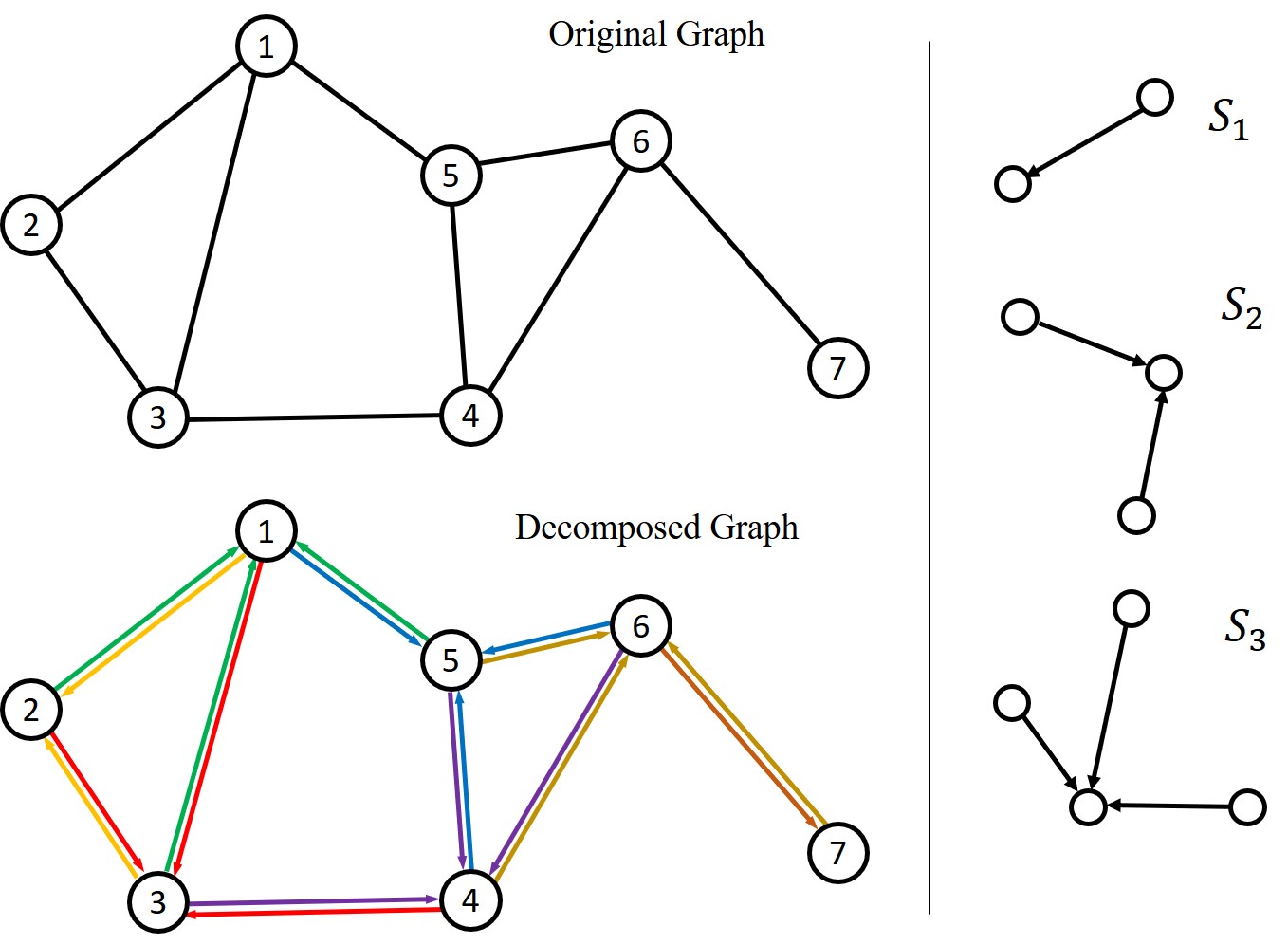}
	\caption{The original graph is decomposed as $\mathcal{G} = \bigcup_{i=1}^{7} \mathcal{S}_{d_i}$, where $d_1=d_3=d_4=d_5=d_6=3$, $d_2=2$, and $d_7=1$. }
	\label{fig:star_graph}
\end{figure}

\subsection{Graph Decomposition}
In this work, we assume that all communication graphs are undirected \cite{van2010graph}, i.e., every communication link is bidirectional. Every undirected graph with $N$ nodes and maximum degree $\delta$ can be decomposed into $N$ star subgraphs in the following sense. Every given undirected graph $\mathcal{G}$ with $N$ nodes and node degrees $d_1, \ldots, d_N$ can be decomposed into $N$ directed star graphs $\mathcal{S}_{d_i}$ such that 
\begin{equation} \label{graph-decom}
\mathcal{G} = \bigcup_{i=1}^{N} \mathcal{S}_{d_i},
\end{equation}
where $\mathcal{S}_{d_i}$ represents a star graph with $d_i$ directional edges that all of them are pointing towards the center node. We refer to Fig. \ref{fig:star_graph} for an illustration. It is straightforward to verify that this decomposition always exists and easy to find as every undirected edge can be decomposed into two directional edges with opposite directions, i.e., $\{i,j\} = \{(i,j)\} \cup \{(j,i)\}$.   

The graph decomposition \eqref{graph-decom} can be utilized to reduce the offline learning to only star-shaped graphs. The class of all undirected graphs with $N$ nodes whose maximum node degree is less than or equal to $\delta$ can be constructed using the union of a subset of directed star graphs $\{\mathcal{S}_1, \mathcal{S}_{2}, \ldots, \mathcal{S}_{\delta}\}$. This has an important practical implication: one needs to train the robots w.r.t the class of directed star communication graphs $\{\mathcal{S}_1,\ldots, \mathcal{S}_{\delta}\}$. Since $\delta \leq N$, there are at most $n$ different star graphs during training.

\begin{remark}
The design parameter $\delta$ is determined in the training stage. When robots are deployed in an environment, based on their communication range, some may have more than $\delta$ neighbors that they can communicate with. In such situations, at each time step, we allow robots to choose only $\delta$ of their neighbors uniformly at random for communication. This will guarantee that the maximum node degree of the underlying communication graph will not exceed $\delta$ for all time.    
\end{remark}

\subsection{Achieving Localized  Learning via Graph Decomposition}

For $N$ robots, there are $\binom{N}{2} = \frac{1}{2}N(N-1)$ distinct pairs of nodes, where each pair determines one possible communication link between the two robots. Therefore, the number of possible graphs without loops or multiple links is $2^{\binom{N}{2}}$. Instead of aiming to train our robots w.r.t all possible communication graphs, which grows exponentially with $n$, we utilize our proposed graph decomposition \eqref{graph-decom} and train our robots w.r.t only star graphs $\{\mathcal{S}_1, \ldots, \mathcal{S}_{\delta}\}$ for a given $\delta$. This procedure will significantly reduce the computational cost of (offline) training because by fixing the design parameter $\delta$, the computational cost will remain the same for all graphs with maximum degree less than or equal to $\delta$. Thus, the training computational cost becomes independent of $N$. In a nutshell, instead of training robots w.r.t all possible communication graphs, we train each robot to learn how to communicate with at most $\delta$ neighboring robots.        

Our approach is particularly useful for map classification tasks over large spatial domains using a large network of communicating robots with localized sensing capabilities.  


\section{Data Fusion and Learning Algorithm}
In the previous section, we explained how we can achieve scalability using the graph decomposition. In the following, we discuss how each robot is trained to perform the map classification task.   
\subsection{Feature Extraction from Environment}
The diagram of the distributed map classification architecture is shown in Fig. \ref{fig:diagram}. The $i$'th robot's observation at time $t$ is denoted by $O(i,t)$ and its spatial position by ${l}(i,t) \in \mathbb{R}^{2}$. The incoming visual observation is fed into a pre-trained VGG-19 \cite{simonyan2014very} model that is represented  by 
\begin{equation} \label{equ:vgg}
     {v}(i,t) = V \big( O(i,t) \big),
\end{equation}
in which $ {v}(i,t) \in \mathbb{R}^a$ contains relevant information for classification purposes only. Robots are also equipped with a CNN-based goal processor as proposed in \cite{mousavi2019layered}. This unit generates relevant features to identify the most informative location on the map so that the robot can be steered towards that location in the next steps. The goal feature vector is shown by
\begin{equation} \label{equ:goal_based}
     {u}(i,t) = G_1 \big( O(i,t),  {l}(i,t) \big),
\end{equation}
in which $ {u}(i,t) \in \mathbb{R}^{b}$ contains relevant information for a  goal-based motion planning. The output vectors of these two processors are represented as $ {x}(i,t) = \big[ {v}(i,t)^T,  {u}(i,t)^T \big]^T$. 

\subsection{Encoding Feature History with LSTM}
Robots should save and learn relevant connections among their past observations because they can only observe their surrounding environment at each time step. Saving every feature vector $x(i,t)$ is straightforward, but not efficient for communication. In this case, there are two ways of communication. A robot either shares the full history $\{x(i,1),\dots,x(i,t)\}$ or only shares $x(i,t)$. If it shares the full history, the overall message traffic will increase as time goes by, and it will eventually become a burden for the communication system. On the other hand, sharing only $x(i,t)$ is impractical since the communication topology is time-varying; the other robots may not receive enough information. To overcome this challenge, we use a Long-Short Term Memory (LSTM) \cite{lstm} cell to efficiently store the history of features and learn the interrelations among the partial features. All robots are equipped with identical LSTM cells. The dynamics of the LSTM cell is governed by  
\begin{equation} \label{equ:lstm}
    \big[ {m}(i,t)^T,  {w}(i,t)^T \big]^T  = F \big( \; {m}(i,t-1),  {w}(i,t-1),  {x}(i,t) \; \big),
\end{equation}
where $m(i,t)$ (output state) contains the history of the past features observed by robot $i$ up to time $t$, ${w}(i,t)$ is the cell state of LSTM. For communication purposes, robots exchange $m(i,t)$ over the network, where this vector has a fixed size and it contains the full feature history of a robot.


\subsection{Communication and Information Fusion}  \label{sec:com}

\begin{figure}[t]
	\centering
	\includegraphics[width=\linewidth]{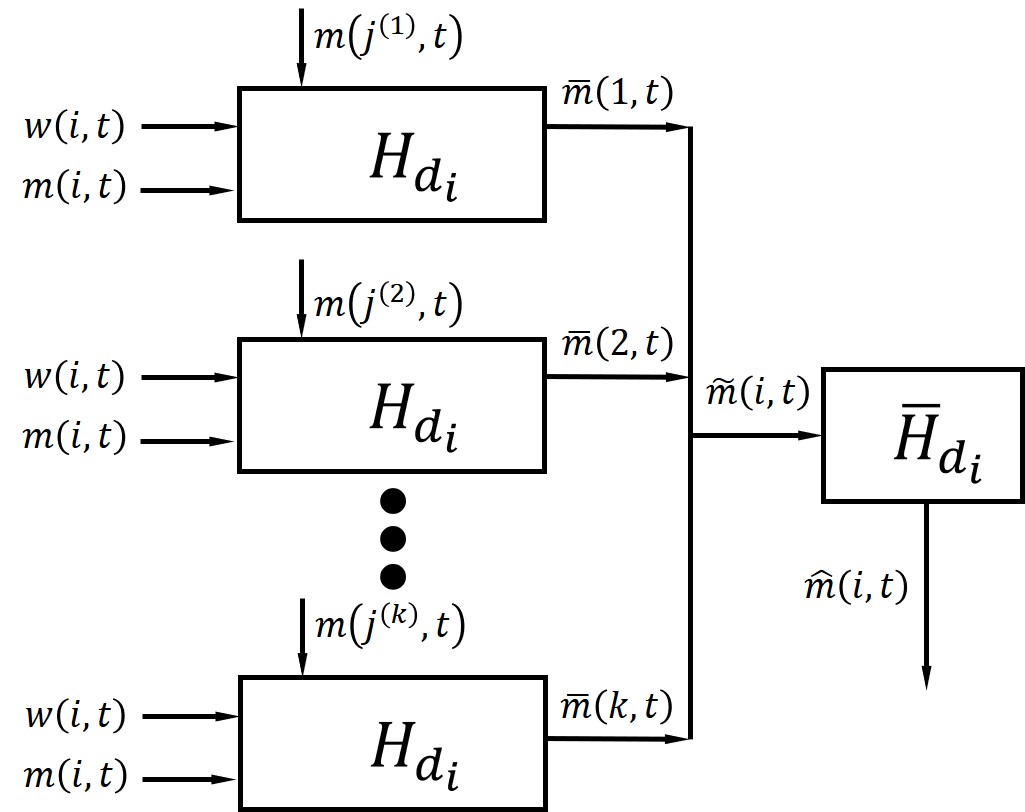}
	\caption{The block diagram of the message processor selected for robot $i$ based on its corresponding star graph $\mathcal{S}_{d_i}$. The neighboring robots of robot $i$ in $\mathcal{S}_{d_i}$ are tagged by  $\{j^{(1)},\dots,j^{(k)}\}$.}
	\label{fig:com_structure}
\end{figure}

A unique message processor is trained for every star graph $\mathcal{S}_k$, where $k=1,\ldots,\delta$. The message processors should be designed with the following capabilities. First, they should be able to communicate and fuse messages received from up to $\delta$ neighbors with their own history ${m}(i,t)$. Second, the output of the message processor, which is denoted by $\hat{{m}}(i,t)$, should be permutation invariant, i.e., it should not depend on the order of messages received from its neighbors. A message processor with these two properties for robot $i$, whose corresponding star graph is $\mathcal{S}_{d_i}$, can be realized by utilizing ${d_i}$ parallel identical LSTM cells $H_{d_i}$ and a fully connected linear layer $\bar{H}_{d_i}$. We refer to Fig. \ref{fig:com_structure} for a block diagram of the proposed structure for this message processor. The $j$'th LSTM cell $\bar{H}_{d_i}$ in the block diagram fuses messages received from the $j$'th neighbor of robot $i$. Moreover, since a linear layer $\bar{H}_{d_i}$ is employed to fuse outputs of the LSTM cells, the final output of the message processor, which is represented by $\hat{{m}}(i,t)$, is always permutation invariant. For every $1 \leq k \leq \delta$, we train and obtain sub-blocks $H_k$ and $\bar{H}_k$ offline and save them in a look-up library for the online implementation.  

The pseudo code for the proposed communication structure is shown in Algorithm \ref{alg:mp}. All robots run this communication mechanism  synchronously to exchange their information: 
\begin{itemize}
    \item \underline{Lines 1-2:} A robot sends a message to its neighbors and receives up to $\delta$ messages from its own neighbors at each time step. If robot $i$ receives $d_i$ messages, then $H_{d_i}$ and $\bar{H}_{d_i}$ will be selected from the look-up library to construct its corresponding message processor. 
    \item \underline{Lines 3-5:} Each parallel LSTM cell $H_{d_i}$ fuses incoming message ${m}(j,t)$ from the $j'$th neighbor of robot $i$ with $ {m}(i,t)$ and ${w}(i,t)$.
    \item \underline{Lines 6-7:} The outputs of all LSTM cells are fused using a linear map $\bar{H}_{d_i}$ to generate the overall output $\hat{{m}}(i,t)$.
\end{itemize}

By omitting some of the details, we highlight that the output $\hat{{m}}(i,t) = \big[\hat{v}(i,t)^T, \hat{u}(i,t)^T \big]^T$ has the same dimension and structure as those of ${m}(i,t)$. This output vector contains the fused information of features for both classification and goal-based motion planning. Each robot sends $\hat{v}(i,t)$ and $\hat{u}(i,t)$ separately to the classifier and goal-based motion planner. 


\begin{algorithm}[t] 
    \KwData{${m}(i,t),  {w}(i,t)$, and $ {m}(j,t)$ from neighbors}
    \KwResult{$\hat{{m}}(i,t)$ }
    Decompose $\mathcal{G}(t)$ as $\bigcup_{i=1}^{N} \mathcal{S}_{d_i}$\;
    Categorize $\mathcal{S}_{d_i}$, and select $H_{d_i}$, and $\bar{H}_{d_i}$\;
        \While{$j \in \mathcal{S}_{d_i}$}{
        $\bar{{m}}(j,t) \leftarrow H_{d_i} \big(  {m} _{i,t},  {w} _{i,t},  {m}(j,t) \big)$\;
        $j \leftarrow j + 1$\;
        }
    $ {\tilde{m}}(i,t) \leftarrow [\bar{ {m}}(1,t)^T, \ldots, \bar{{m}}({d_i},t)^T ]^T$\;
    $\hat{ {m}}(i,t) \leftarrow \bar{H}_{d_i} \big(  {\tilde{m}}(i,t) \big)$\;
 \caption{Communication Structure}
 \label{alg:mp}
\end{algorithm}

\subsection{Motion Planning and Map Classification Tasks}  \label{sec:gmp}

Robots use a goal-based motion planner \cite{mousavi2019layered} to plan their motion with the fused goal feature vector $\hat{u}(i,t)$. They will be navigated toward the most informative region in the map, which is denoted by the goal position ${g}(i,t) \in \mathbb{R}^{2}$. The goal location is sampled as
\begin{align*}
     {g}(i,t) = G_2 \big( \hat{  u}(i,t) \big).
\end{align*}

Meanwhile, robots attempt to classify the target environment with the fused feature history vector $\hat{v} (i,t)$. The classifier consists of fully connected layers followed by a SoftMax function. It generates a prediction vector $q(i,t) \in \mathbb{R}^{M}$ by
\begin{equation*}
      q(i,t) = C \big( \hat{v} (i,t) \big).
\end{equation*}
The predicted label $P$ is generated with a Argmax function such that $P = \arg \max \big( q(i,T) \big)$.

For a dataset that consists of $M$ labels, the ground truth vector for the $k'$th class is shown by ${Q}_k$, which is defined as the $k'$th Euclidean basis in $\mathbb{R}^{M}$. The reward for classifying the $k'$th label with robot $i$ is evaluated by a log-sum-exp (LSE) loss
\begin{equation*}    \label{equ:lse_loss}
    r(i,T) = -LSE \big( {Q}_k,  {q}(i,T) \big). 
\end{equation*}

The pseudo code of distributed map classification is presented in Algorithm \ref{alg:full}. At the beginning, we initialize the location of robots $l(i,0)$, and reset feature history $m(i,0)$ and cell state $w(i,0)$ to zero for all robots. 
\begin{itemize}
    \item \underline{Lines 4-5:} Robots take localized observations, process it with image and goal processors and encode the feature history with the LSTM unit.
    \item \underline{Lines 9-11:} Robots communicate with their neighbors and sample the goal location for navigation. 
    \item \underline{Lines 16-17:} Robots classify the map at the end of the task.
\end{itemize}  

\begin{algorithm}[t] 
    \KwResult{Predicted Label $P$ }
        Initialize $ {l}(i,0)$, $O(i,0)$, $ {m}(i,0)$ and $ {w}(i,0)$ for all robots\;
        \While{$t \leq T$, }{
            \While{$i \in \mathcal{V}$}{
                update $ {x}(i,t)$ with Equ. \eqref{equ:vgg} and \eqref{equ:goal_based}\;
                update feature history ${m}(i,t)$ with Equ. \eqref{equ:lstm}\;
                $i \leftarrow i + 1$\;
            }
            \While{$i \in \mathcal{V}$}{
                $[\hat{ {v}}(i,t)^T, \hat{ {u}}(i,t)^T]^T \leftarrow$ \textbf{Algorithm \ref{alg:mp}}\;
                $ {g}(i,t) \leftarrow G_2 \big( \hat{  u}(i,t) \big)$\;
                update $ {l}(i,t)$ w.r.t. goal location\;
                $i \leftarrow i + 1$\;
            }
        $t \leftarrow t + 1$\;    
    }
    $q(i,T) \leftarrow C \big( \hat{  v} (i,T) \big)$\;
    $P \leftarrow \arg \max \big( q(i,T) \big)$\;
    \caption{Distributed Multi-robot Map Classification}
    \label{alg:full}
\end{algorithm}


\section{Case Studies and Simulations}      \label{sec:case}

\subsection{Satellite Map Dataset}

\begin{figure}
	\begin{subfigure}[t]{.31\linewidth}
		\centering
		\includegraphics[width=\linewidth]{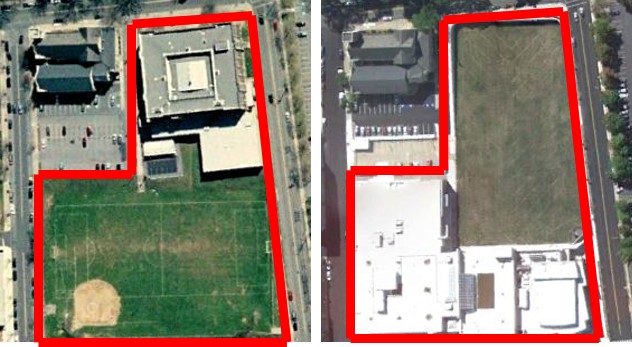}
		\caption{Long-term feature changes (years).}
	\end{subfigure}
	\hfill
	\begin{subfigure}[t]{.35\linewidth}
		\centering
		\includegraphics[width=\linewidth]{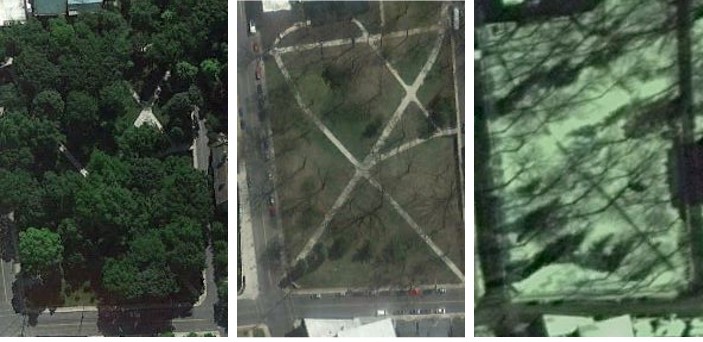}
		\caption{Short-term features change (seasons).}
	\end{subfigure}
	\hfill
	\begin{subfigure}[t]{.27\linewidth}
		\centering
		\includegraphics[width=\linewidth]{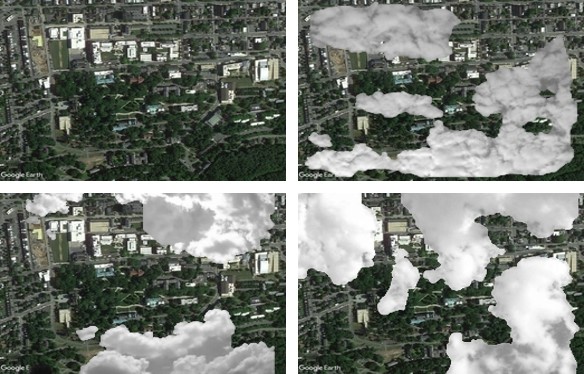}
		\caption{Maps with and without clouds.}
	\end{subfigure}
	\caption{Satellite map dataset.}
	\label{fig:campus}
\end{figure}

A map dataset is created by using satellite maps exported from Google Earth of 10 university campuses over the past 40 years\footnote{All maps and figures used in this work are exported from Google Earth. They have been customized by the author and do not indicate what it appears online in Google Earth.}, we refer to Fig. \ref{fig:campus}. To simulate the real-world environment, we added 80 randomly generated clouds in each map of the original dataset, such that each cloud will cover $40\%$ of the map (Fig.\ref{fig:campus}(c)). Such manipulation will increase the difficulty of the classification since 1) the total available area of the map is constrained; 2) different maps now share more similar features (clouds). The clouded map dataset consists of 20,000 training maps and 2,000 unseen testing maps. To successfully classify the map, robots need to filter out minor changes and focus only on the major features of the map. 

The size of each map is 1024$\times$768 pixels, and robots are allowed to take square observations with a frame size of $p^2 = 64\times64$ pixels (relative observation size 0.52 \%). The communication range is set as 480 pixels to encourage robots to use various star graphs when training the communication. 

\subsection{Simulation on MNIST and Map Dataset}
The estimated cost function of robot $i$ is defined as $J_i = r(i, T)$. Implementation with RL refers to solve the optimization problem of maximizing $J_i$ subject to constructing $\tilde{\mathcal{G}}(t)$ with star graphs, a finite number of robots $N$ and finite time horizon $T$. To ensure the optimal performance among all robots, we replace $J_i$ with a global average. 
$J = (\sum_{i=1}^{N} r(i, T))/N$. All robots are equipped with identical learned models such that the updated hyperparameters will be applied to all robots after each training epoch. 

\begin{figure}[t]
	\includegraphics[width=\linewidth]{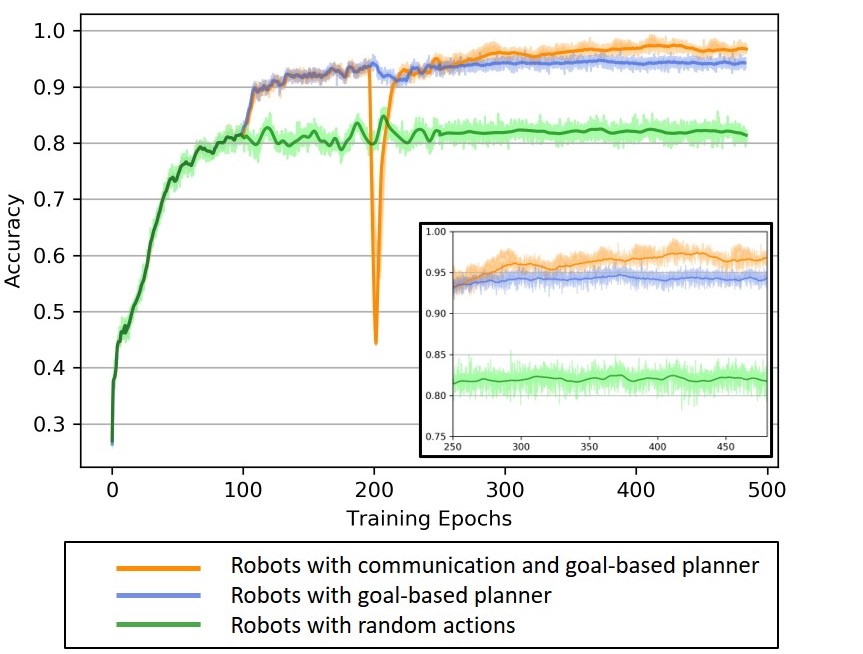}
	\caption{The learning curves with magnified details. Robots with goal-based motion planners achieve a better performance than random actions as they explore the most informative region. The drop in the orange curve shows learning the communication policy from scratch at 200 epoch.}
	\label{fig:compare}
\end{figure}

\begin{figure*}[t]
	\begin{center}
		\makebox[\textwidth]{\includegraphics[width=\linewidth]{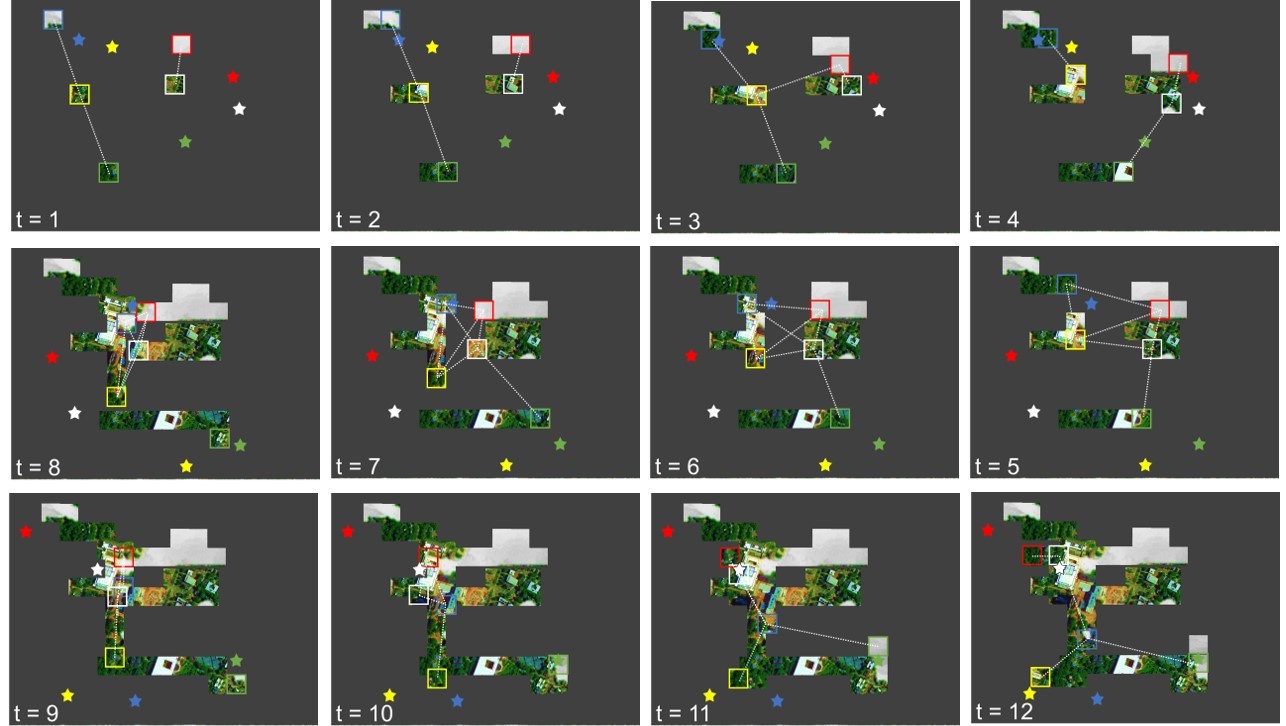}}
		\caption{The snapshots taken from experiments with $N = 5$ and $T = 12$. The observations and locations of robots are shown by the colored square. Correspondingly, the goal locations are denoted by stars, and the communication links are denoted by dashed lines.}
		\label{fig:step}
	\end{center}
\end{figure*}

We use ADAM \cite{kingma2014adam} with a learning rate $l_r = 0.0001$ to train the model in PyTorch \cite{paszke2017automatic}. As shown in Fig. \ref{fig:compare}, to ensure the robots can learn the communication policies, we only enable the training on communication after robots have been trained to the best performance independently. The accuracy is presented as a global average among all robots and five random seeds. The method is validated over both the satellite map dataset and the MNIST dataset \cite{lecun1998gradient}. Table. \ref{table:accuracy} indicates the proposed distributed method achieves a comparable performance w.r.t. the centralized method, in which we train a VGG-19 model with the entire image as input. In Fig. \ref{fig:step}, we show the snapshots from an experiment with $N=5$ and $T = 12$.

\begin{table}[t]                           
	\centering	
	\resizebox{\linewidth}{!}{ 
		\begin{tabular}{l|c|ccc|c}
			\toprule
			Dataset	& Observation	&	1 robot	&	5 robots	&	10 robots  &  VGG-19 \\
				& size (\%)	&	 	&	 	&	  &  w/ full image \\
			\midrule
			MNIST & 2.04 & 91.27 & 94.98 & 98.31 &  99.33             \\ 
			\midrule
			Map  &   0.52 & 85.19 & 98.94 & 99.71 &  99.84         \\ 
			w/o clouds &   2.08 & 88.59 & 99.27 & 99.84 &  99.84         \\  
			\midrule
			Map &   0.52 & 72.42 & 97.30 & 97.56 & 99.43          \\   
			w/ clouds &   2.08 & 77.62 & 98.21 & 98.98 & 99.43          \\   
			\toprule
	\end{tabular} }                        
	\caption{The classification accuracy (\%) over the MNIST and the satellite map dataset.}
	\label{table:accuracy}  
\end{table}

\subsection{Scalable Learning with Star Graph Decomposition} 

To validate the scalable learning, two models are trained using a complete communication graph (all to all) and our proposed method with $\delta = 4$. Both models are trained with $N = 5$, and $T = 15$ on the clouded map dataset. As shown in Table. \ref{table:scale}, in the case of using the star graph decomposition (Star-<5>), it preserves a high performance when new robots join the network without additional training. However, in the case of using the complete graph (Complete-<5>), the performance will decrease as the number of robots changes. Besides, we test the scalability of the complete communication network in the other direction. We trained the complete graph with $N = 80$ (Complete-<80>), and it does not preserve the scalability vice versa. 

Using the star graph decomposition reduces the training cost significantly for a large number of robots. We evaluate the extra training time used for Complete-<5> to reach the same performance level of Star-<5> as in the first row of Table. \ref{table:scale}. The result in Table. \ref{tab:time} indicates once the number of robots increases, the extra training time need for the model with a complete communication network increases drastically. Using star graph decomposition can save a significant amount of time since it is trained with only a few robots ($N \geq \delta$), and it preserves the high performance with a large number of robots.

\begin{table}[t]                           
	\centering	
	\resizebox{\linewidth}{!}{ 
		\begin{tabular}{l|ccccc}
			\toprule
			Method $\&$ Number of robots &	5	&	10	&	20	&	40  &  80 \\
			\midrule
			Star-<5>	&   97.30 & 97.56 & 96.90 & 97.20 &  96.72             \\ 
			Complete-<5>	&   95.60 & 93.21 & 86.55 & 85.46 &  63.36          \\  
			Complete-<80>&   73.54 & 76.81 & 78.55 & 90.59 & 98.24          \\  
			\toprule
	\end{tabular} }                        
	\caption{The classification accuracy (\%) tested with the various number of robots. All the data is evaluated without additional training.}
	\label{table:scale}   
\end{table}

\begin{table}[t]                           
	\centering	
	\resizebox{\linewidth}{!}{ 
		\begin{tabular}{l|ccccc}
			\toprule
			Number of Robots	&	5	&	10	&	20	&	40 & 80 \\
			\midrule
			Additional training time (min)   &   $\infty$ & 320  & 1132  & 1248  & 2945 \\  
			\toprule
	\end{tabular} }                        
	\caption{The additional training time used for a complete communication network to reach the same performance as Star-<5>, evaluated on an NVIDIA Tesla K80 GPU.}
	\label{tab:time}   
\end{table}

\begin{figure}[t]
	\includegraphics[width=\linewidth]{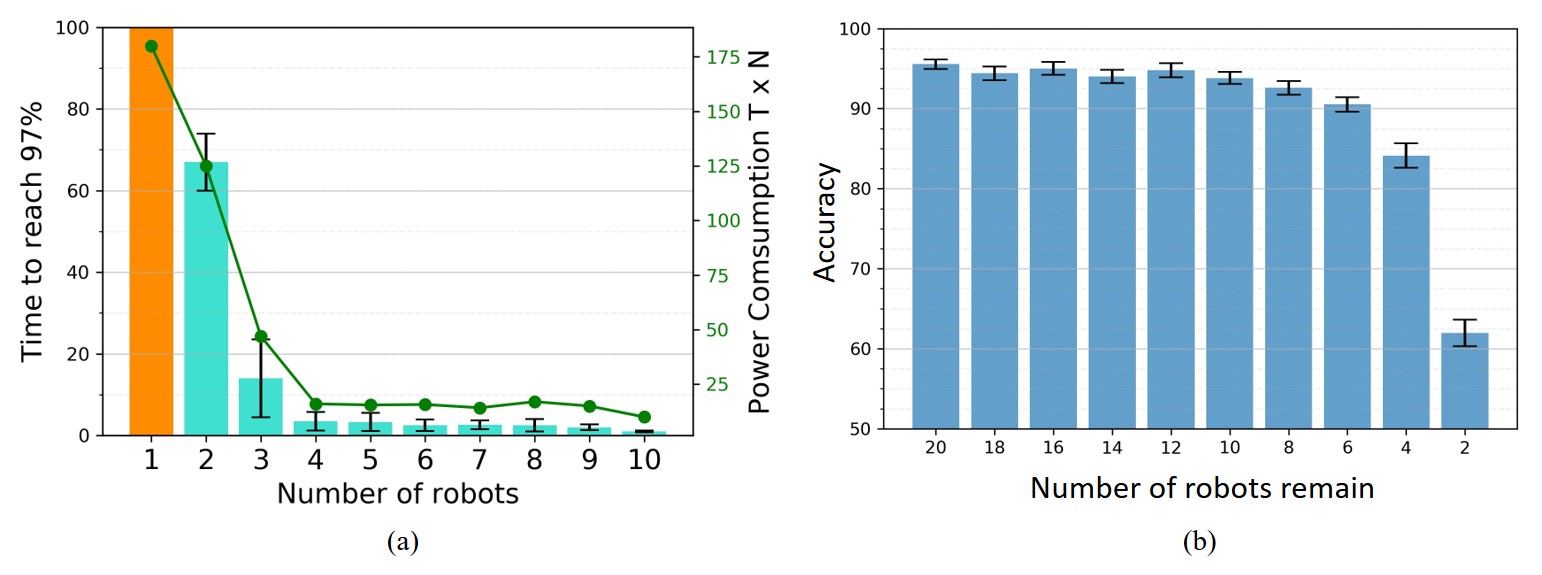}
	\caption{(a): The average time used to reach classification accuracy of $97\%$ with a various number of robots. The orange bar indicates it takes approximately $T=180$ for a single robot to complete the task. (b): The performance (\%) after removing $\{ 2,...,18 \}$ robots from the team.}
	\label{fig:time}
\end{figure}

\subsection{Efficiency and Robustness of Distributed Classification}

We evaluate the average time used for robots to reach a classification accuracy threshold ($97\%$) in Fig. \ref{fig:time} (a). It shows an intuitive result that to achieve a certain accuracy, the time cost $T$ is inversely proportional to the number of robots $N$. Since all robots are identical, the total power consumption (e.g., battery life) $T \times N$ is a valid measure for the cost of the completion of a task. We refer to the result in Fig. \ref{fig:time} (a), which indicates the total energy cost for a network of robots would be dramatically reduced as the number of robots increases. It should be emphasized that there exists an optimal number of robots for a certain task since it takes at least $T=1$ for any number of robots to complete the task.

The star graph decomposition grants the robustness to the architecture since robots can always reform a new communication network $\tilde{\mathcal{G}}(t)$ with star graphs $\{\mathcal{S}_1,\dots, \mathcal{S}_{\delta} \}$, which they have been trained with. We validate the robustness by letting 20 robots trained with Star-<5> classify the map with $T = 10$. We randomly remove some robots from the network during the task and evaluate the remaining robots' testing results. The result is shown in Fig. \ref{fig:time} (b) as the proposed architecture preserves a good performance until losing $80\%$ of the robots.  Such decomposition has many other practical implications, including minimizing the number of message transmissions over the network and enhancing network privacy and security due to short-range communication. 

\section{Conclusion} 
We present a distributed multi-robot classification architecture that allows a large network of robots with localized sensing capabilities to classify the map. The utilization of the graph decomposition allows the robots to decompose the time-varying communication graph, and learn the communication in a localized and scalable manner. We demonstrate the usefulness of the proposed method by using multi-robot image and map classification as an example. Our proposed distributed method can achieve a high classification performance with only partial observations compared to the centralized methods. Besides, the graph decomposition enables the possibility of localized and scalable learning for the architecture. It significantly reduces the training cost as one can learn the model with a relatively small number of robots. With such a model, the performance is still maintained in the case of a large number of robots and the time-varying topology without additional training cost. The utilization of a distributed multi-robot system also saves the energy for complete a specific task and provides robustness to the system since it is exceptionally compatible with the time-varying topology.







\printbibliography

\end{document}